\documentclass[sigconf]{acmart}
\usepackage{enumitem}
\usepackage{bbm}
\usepackage{multirow}
\usepackage{caption}
\usepackage{graphicx}
\usepackage{float} 
\usepackage{subcaption}
\usepackage{balance}
\usepackage[linesnumbered, ruled]{algorithm2e}
\SetKwRepeat{Do}{do}{while}%
\AtBeginDocument{%
  \providecommand\BibTeX{{%
    \normalfont B\kern-0.5em{\scshape i\kern-0.25em b}\kern-0.8em\TeX}}}

\setcopyright{acmcopyright}
\copyrightyear{2023}
\acmYear{2023}
\setcopyright{acmlicensed}\acmConference[CIKM '23]{Proceedings of the 32nd
ACM International Conference on Information and Knowledge
Management}{October 21--25, 2023}{Birmingham, United Kingdom}
\acmBooktitle{Proceedings of the 32nd ACM International Conference on Information and Knowledge Management (CIKM '23), October 21--25, 2023,
Birmingham, United Kingdom}
\acmPrice{15.00}
\acmDOI{10.1145/3583780.3615023}
\acmISBN{979-8-4007-0124-5/23/10}

\begin{document}

\title{RDGSL: Dynamic Graph Representation Learning with Structure Learning}

\author{Siwei Zhang}
  \email{swzhang22@m.fudan.edu.cn}
  \affiliation{%
    \institution{Shanghai Key Laboratory of Data Science, School of Computer Science, Fudan University}
    \city{Shanghai}
    \country{China}
   }

\author{Yun Xiong}
\authornote{Corresponding author}
\author{Yao Zhang}
 \email{yunx@fudan.edu.cn}
 \email{yaozhang@fudan.edu.cn}
  \affiliation{%
     \institution{Shanghai Key Laboratory of Data Science, School of Computer Science, Fudan University}
     \city{Shanghai}
    \country{China}
  }

\author{Yiheng Sun}
  \email{sunyihengcn@gmail.com}
  \affiliation{%
    \institution{Tencent Weixin Group}
    \city{Shenzhen}
    \country{China}
  }

\author{Xi Chen}
  \email{x_chen21@m.fudan.edu.cn}
  \affiliation{%
    \institution{Shanghai Key Laboratory of Data Science, School of Computer Science, Fudan University}
    \city{Shanghai}
    \country{China}
  }

\author{Yizhu Jiao}
  \email{yizhuj2@illinois.edu}
  \affiliation{%
    \institution{University of Illinois at Urbana-Champaign}
    \state{IL}
    \country{USA}
  }

\author{Yangyong Zhu}
 \email{yyzhu@fudan.edu.cn}
  \affiliation{%
     \institution{Shanghai Key Laboratory of Data Science, School of Computer Science, Fudan University}
     \city{Shanghai}
    \country{China}
  }

\renewcommand{\shortauthors}{Siwei Zhang, et al.}

\newcommand \footnoteONLYtext[1]
{
	\let \mybackup \thefootnote
	\let \thefootnote \relax
	\footnotetext{#1}
	\let \thefootnote \mybackup
	\let \mybackup \imareallyundefinedcommand
}

\begin{abstract}
Temporal Graph Networks (TGNs) have shown remarkable performance in learning representation for continuous-time dynamic graphs. However, real-world dynamic graphs typically contain diverse and intricate noise. Noise can significantly degrade the quality of representation generation, impeding the effectiveness of TGNs in downstream tasks. Though structure learning is widely applied to mitigate noise in static graphs, its adaptation to dynamic graph settings poses two significant challenges. i) \textbf{Noise dynamics.} Existing structure learning methods are ill-equipped to address the temporal aspect of noise, hampering their effectiveness in such dynamic and ever-changing noise patterns. ii) \textbf{More severe noise.} Noise may be introduced along with multiple interactions between two nodes, leading to the re-pollution of these nodes and consequently causing more severe noise compared to static graphs.

In this paper, we present \textbf{RDGSL}, a representation learning method in continuous-time dynamic graphs. Meanwhile, we propose dynamic graph structure learning, a novel supervisory signal that empowers RDGSL with the ability to effectively combat noise in dynamic graphs. To address the noise dynamics issue, we introduce the Dynamic Graph Filter, where we innovatively propose a dynamic noise function that dynamically captures both current and historical noise, enabling us to assess the temporal aspect of noise and generate a denoised graph. We further propose the Temporal Embedding Learner to tackle the challenge of more severe noise, which utilizes an attention mechanism to selectively turn a blind eye to noisy edges and hence focus on normal edges, enhancing the expressiveness for representation generation that remains resilient to noise. Our method demonstrates robustness towards downstream tasks, resulting in up to 5.1\% absolute AUC improvement in evolving classification versus the second-best baseline.
\end{abstract}




%%  \includegraphics[width=\textwidth]{sampleteaser}
%%  \caption{Seattle Mariners at Spring Training, 2010.}
%%  \Description{Enjoying the baseball game from the third-base
%%  seats. Ichiro Suzuki preparing to bat.}

\maketitle

\section{Introduction}
\label{sec:intro}
In recent years, the exploration of graph representation learning has emerged as a pivotal research area \cite{CATCHM2023, deepMincut2023, graph_1_2022, graph_2_2022, graph_3_2022, graph_4_2023}. In particular, some graphs exhibit dynamic changes in their structure over continuous time points such as social networks \cite{socialNetwork2023} and online shopping networks \cite{DTCG_2_2022}, and this type of graph is commonly referred to as continuous-time dynamic (or temporal) graph~\footnote{For simplicity, we use ``dynamic graph'' in the following of our paper.} \cite{dynamicGraph2022}. To address the problem of representation learning in such graphs, Temporal Graph Networks (TGNs) \cite{CTDNE2018, JODIE2019, DyRep2019, TGAT2020, TGN2020} have been introduced. These networks employ a memory module to capture the historical behaviors of nodes, enabling us to leverage their past information to make predictions about future activities \cite{TGN2020}.

\begin{figure}
  \centering
  \includegraphics[width=0.9\linewidth]{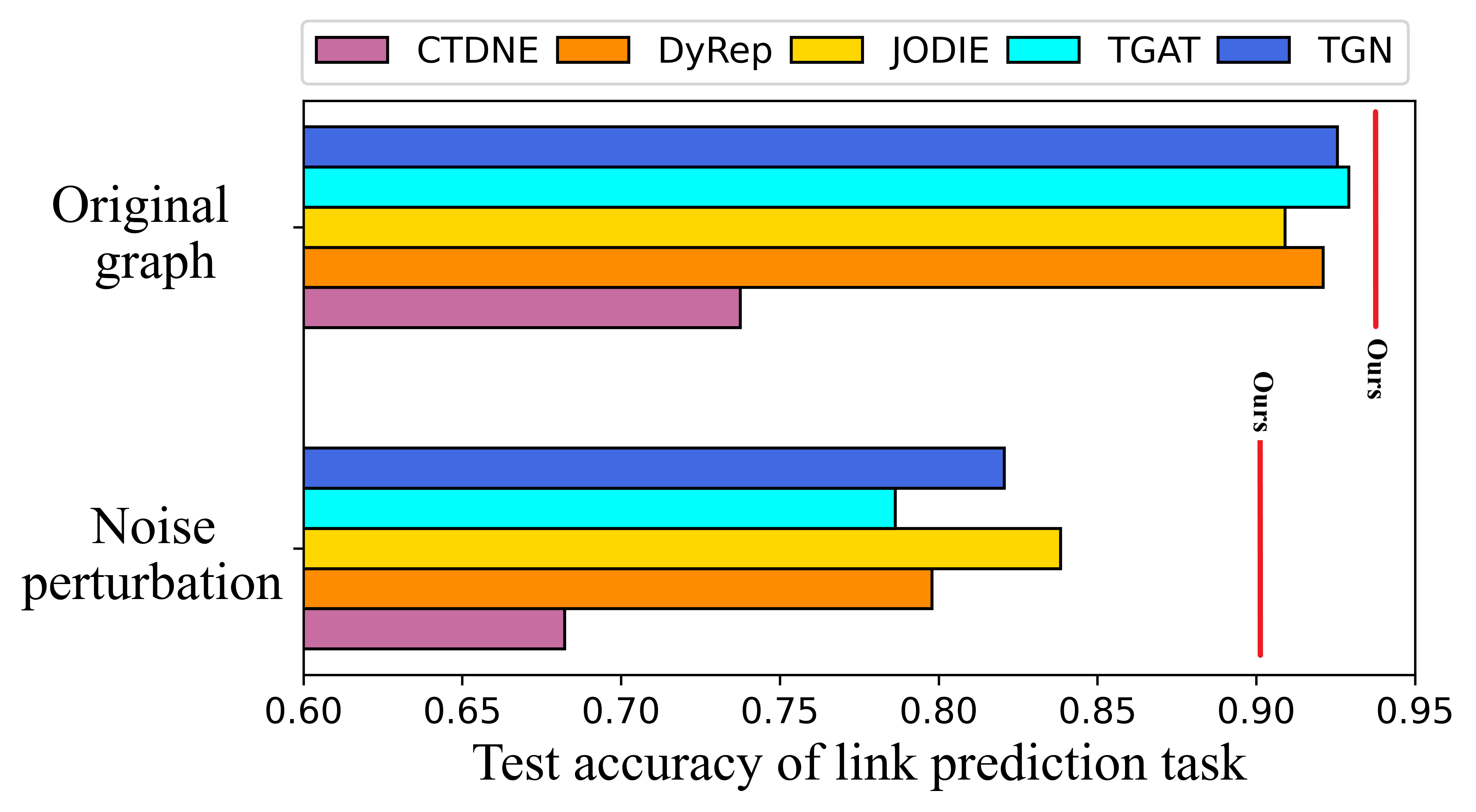}
  \caption{Noise degrades TGNs' performance. We simulate the noise used in \cite{MATA2021} on the Reddit dataset \cite{reddit2020} and conduct link prediction task using TGNs \cite{CTDNE2018, DyRep2019, JODIE2019, TGAT2020, TGN2020}. Note that the red lines represent the performance of our method. For more implementation details, please refer to Section \ref{sec:ExpSetting}.}
  \label{noise}
\end{figure}

\begin{figure}
  \centering
  \begin{subfigure}{\linewidth}
    \centering
    \includegraphics[width=0.9\linewidth]{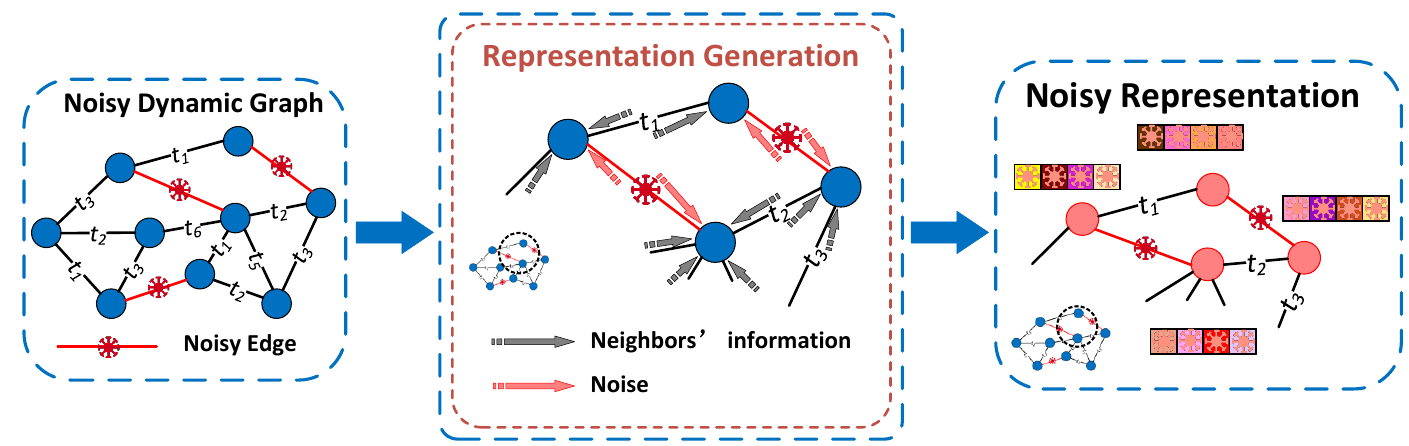}
    \caption{Existing TGNs ignore the noise during representation generation.}
    \label{intro_A}
  \end{subfigure}
  \qquad
  \centering
  \begin{subfigure}{\linewidth}
    \centering
    \includegraphics[width=0.9\linewidth]{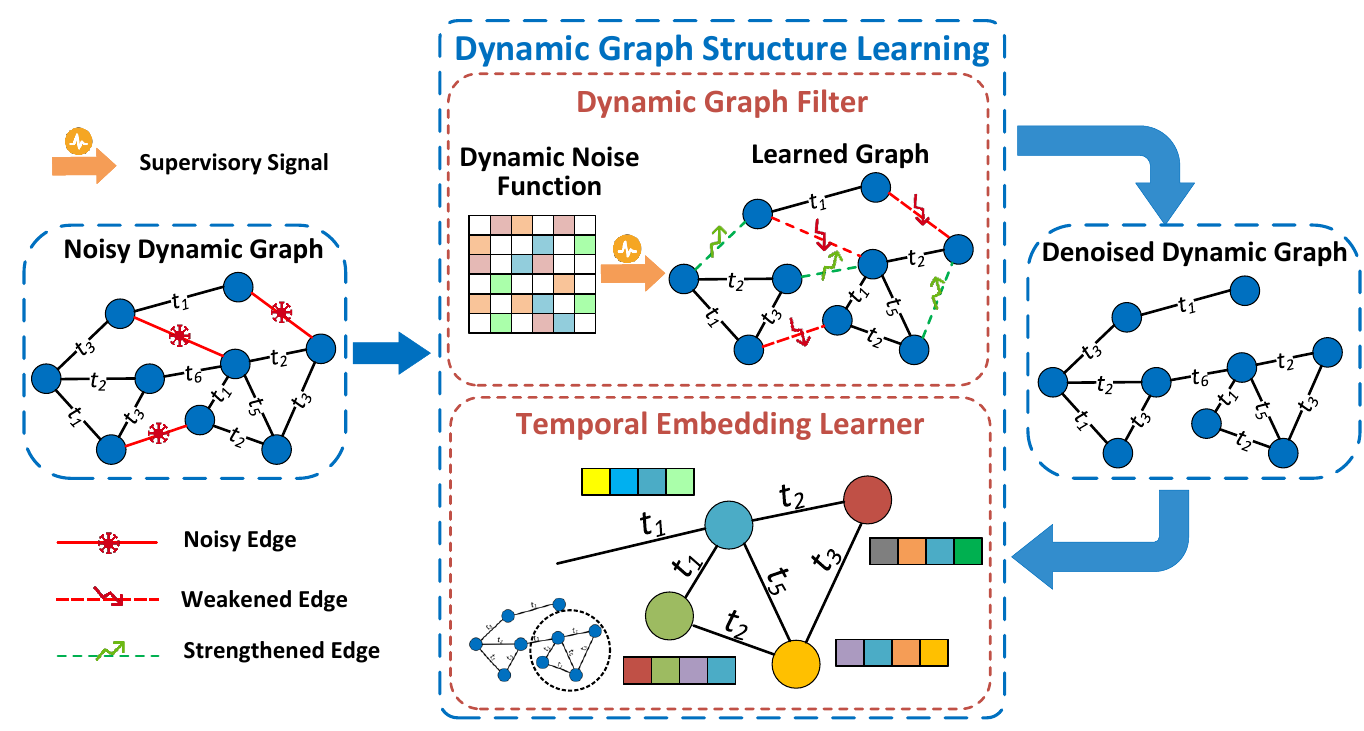}
    \caption{RDGSL can purify noise with structure learning. [This paper]}
    \label{intor_B}
  \end{subfigure}
  \caption{Comparison of the models ignoring and considering the noise in dynamic graphs. (a) Noise will be incorporated and aggregated into representation, reducing TGNs' performance in downstream tasks. (b) With dynamic graph structure learning, our proposed RDGSL can purify noise during the representation generation in dynamic graphs. }
  \label{sensitivity}
\end{figure}

TGNs have demonstrated good performance, but real-world dynamic graphs inevitably contain diverse and complex noise.
We conduct a preliminary experiment on the Reddit dataset \cite{JODIE2019} to test how noise impacts the link prediction results of TGNs in Figure \ref{noise}.
We observe a noticeable reduction in the performance of various existing TGNs due to the presence of noise. Noise has a negative impact on the representation generation process of TGNs, thereby limiting their effectiveness in downstream tasks. Specifically, as depicted in Figure~\ref{intro_A}, when TGNs utilize the message passing mechanism \cite{TGN2020} to aggregate the information of temporal neighbors, the presence of noise will be incorporated and aggregated into the representation, leading to suboptimal outcomes of TGNs \cite{noise2021}. 

Graph Structure Learning (GSL) has attracted considerable attention as an effective strategy for handling noise in static graphs \cite{GSL_1_2022, GSL_2_2022, GSL_3_2022, GSL_0_2020, GSL_4_2022, structure_1_2022}. GSL methods generally consist of two components: a graph generation module and a Graph Neural Network (GNN) embedding module. The graph generation module aims to denoise through a static noise function \cite{similarity2021} that evaluates the noise in static graphs and generates a denoised graph, allowing the GNN module to be optimized in downstream tasks. This naturally raises an intuitive thought: \emph{whether we can explore a structure learning method that effectively handles noise in the representation learning of dynamic graphs.}
\footnoteONLYtext{There can be multiple edges between two nodes in dynamic graphs. For clarity and simplicity, we only depict one edge in all the figures of this paper.}

Despite the effectiveness of GSL methods for denoising in static graphs, their adaptation to dynamic graph settings presents two significant challenges.
i) \textbf{Noise dynamics.}
In dynamic graphs, noise demonstrates remarkable dynamics, characterized by chronologically evolving structure and uncertain variation. GSL methods, which predominantly focus on static graphs, are ill-equipped to address the temporal aspect of noise dynamically. The failure to fully capture and model noise dynamics hampers their effectiveness in handling the noise of dynamic graphs.
ii) \textbf{More severe noise.}
In dynamic graphs, two nodes can interact at different times. Noise in edges may be introduced along with these multiple interactions, leading to the re-pollution of these nodes \cite{MATA2021}. This phenomenon will cause more severe noise compared to static graphs, and consequently, noise reduction task in dynamic graphs becomes increasingly laborious for GSL methods. Therefore, a more effective denoising method tailored to dynamic graphs is imperative.

In this paper, we present \textbf{RDGSL} (\underline{\textbf{R}}epresentation Based on \underline{\textbf{D}}ynamic \underline{\textbf{G}}raph \underline{\textbf{S}}tructure \underline{\textbf{L}}earning), a concrete representation learning method designed to effectively combat noise in continuous-time dynamic graphs. As illustrated in Figure \ref{intor_B}, we propose dynamic graph structure learning, a novel learning method tailored specifically for dynamic graphs. It is a distinctive supervisory signal that aims to weaken the adverse effects of noisy edges while concurrently strengthening the positive impact of normal ones, empowering RDGSL with denoising capabilities in dynamic graphs.
RDGSL comprises two main components:
i) Dynamic Graph Filter. To tackle the noise dynamics issue, we introduce a dynamic noise function, a dual-function module proficient in evaluating temporal noise dynamically. Specifically, the dynamic noise function utilizes a base function to encode and evaluate the current noise for the incoming edge, and a temporal function to dynamically capture and model the noise originating from historical interactions. As a result, a noise-reduced dynamic graph is generated.
ii) Temporal Embedding Learner. To address the challenge of more severe noise, we leverage an attention mechanism that combines the evaluated temporal noise on the noise-reduced dynamic graph to obtain representation. Specifically, we leverage the attention mechanism to selectively turn a blind eye to noisy edges and hence focus on normal edges, which further increases the expressiveness for generating an informative representation that remains resilient to noise. 

In summary, our main contributions are:
\begin{itemize}[leftmargin=*]
\item Our approach constitutes the first attempt to investigate structure learning in continuous-time dynamic graphs, which equips our method to resist the noise in dynamic graphs.
\item We present RDGSL, a concrete method for dynamic graph representation learning. Different from existing TGNs, RDGSL focuses on effectively resisting noise in dynamic graphs.
\item We introduce the dynamic noise function, a dual-function module that dynamically captures the temporal noise in dynamic graphs. Additionally, we combine an attention mechanism with the evaluated temporal noise, further increasing the expressiveness of our method in noisy dynamic graphs.
\item We conduct extensive experiments on real-world datasets to verify the robustness of RDGSL in noisy dynamic graphs.
\end{itemize}

\section{Related Work}
\subsection{Dynamic Graph Representation Learning}
Representation learning on static graphs has gained significant attention recently, but the potential for learning on dynamic graphs remains largely untapped \cite{MATA2021, cope2021, TGN2020}. Early works such as CTDNE \cite{CTDNE2018} focus on generating static node representation by constructing temporal random walks. However, these methods are difficult to apply to new nodes and edges, failing to depict the ever-changing nature of dynamic graphs. Therefore, JODIE \cite{JODIE2019} employs a couple of recurrent networks and the projection operation to generate dynamic representation. DyRep \cite{DyRep2019} and TGAT \cite{TGAT2020} utilize the attention layers to generate dynamic representation by aggregating the information from historical neighbors. TGN \cite{TGN2020} unifies the existing methods by generalizing the model into the memory module, the message-related modules, and the embedding module. PINT \cite{pint2022} introduces a novel method that leverages position features to enhance TGN, achieving state-of-the-art performance in dynamic graph representation learning.

All of these works overlook the presence of noise in real-world dynamic graphs, where inherent noise in original graphs can significantly impair the performance of these methods.

\begin{figure*}
  \centering
  \includegraphics[width=0.95\linewidth]{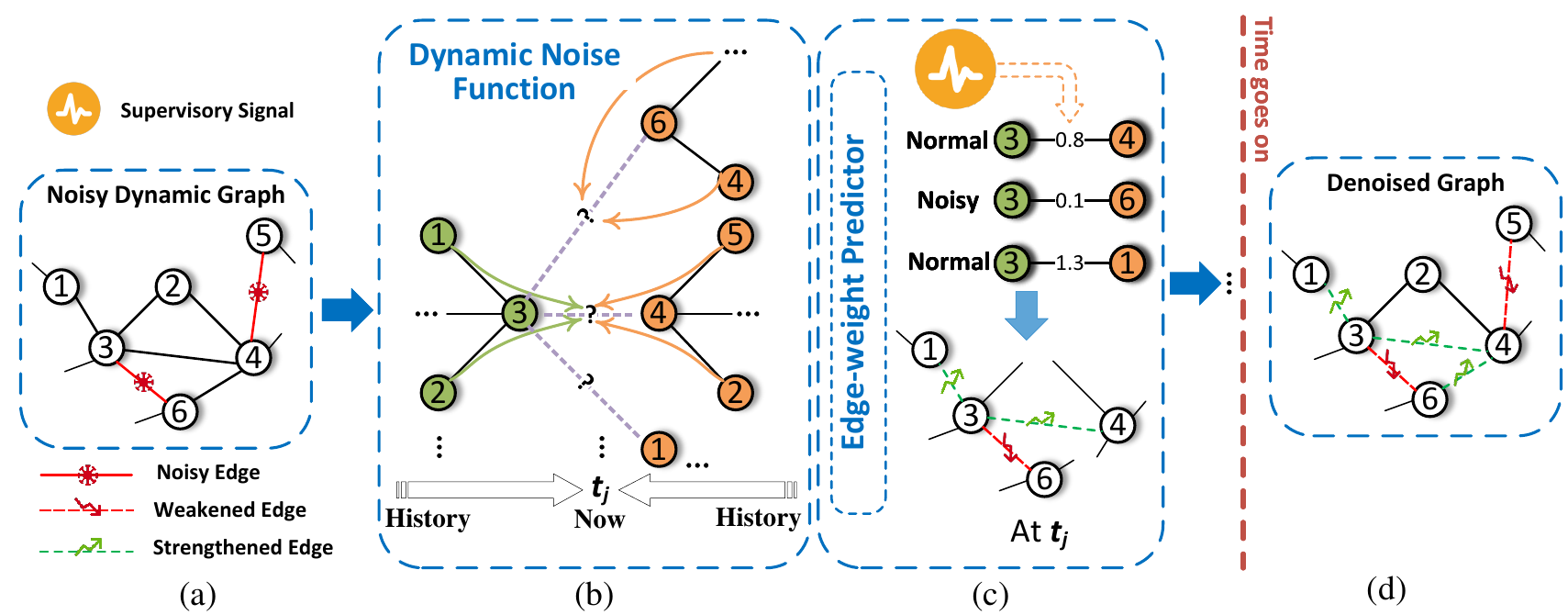}
  \caption{The simplified description of denoising in dynamic graphs. (a) A noisy dynamic graph. Note that we omit the timestamp on edges for convenience. (b) Dynamic noise function. At the current time $t_j$, $v_3$ has three interactions ($v_3v_1, v_3v_4, v_3v_6$ in ``Now''). We take $v_3v_4$ as an example. Our base function (represented by dash lines) is utilized between $v_3$ and $v_4$ to encode the current noise and evaluate its degree at $t_j$. Meanwhile, our temporal function (represented by solid arrows) dynamically captures and models the historical noise originating from historical interactions before $t_j$ ($v_3v_1, v_3v_2$ and $v_4v_5, v_4v_2$ in ``History''), enhancing the ability to assess the temporal aspect of noise. (c) An edge-weight predictor. With the supervisory signal of dynamic graph structure learning, noisy edges ($v_3v_6$) are more likely to get a lower weight, and normal edges ($v_3v_4$ and $v_3v_1$) tend to get a higher one. (d) A denoised dynamic graph. As time goes on, the noisy graph is purified for representation generation. }
  \label{framework}
\end{figure*}

\subsection{Graph Structure Learning (GSL)}
Existing GSL methods primarily consist of two components: a graph generation module and a Graph Neural Network (GNN) module. The purpose of the graph generation module is to eliminate or reduce the noise in static graphs and obtain a denoised structure. Then, the noise-reduced structure is used to optimize the parameter of the GNN module in node classification tasks \cite{GSL_0_2020, GSL_1_2022, GSL_3_2022, GSL_5_2022, GSL_6_2022}. Most GSL methods pay attention to modifying the weights of edges to optimize the specific graph structure, including increasing the weights of normal edges and reducing the weights of noisy edges or eliminating them directly \cite{structure_1_2022, GSL_2_2022, similarity2021}. In general, GSL methods can be roughly divided into three categories \cite{GSLservey2021}: direct optimization method \cite{GSL_0_2020}, probabilistic modeling method \cite{GSL_1_2022}, and metric learning method \cite{RSGNN2022}. The metric learning method is the most widely used recently, which relies on a metric function such as similarity \cite{similarity2021, RSGNN2022} or Gaussian distribution \cite{GSL_1_2022} to learn a weight for every pair of nodes, absorbing the noise in static graphs.

However, existing GSL methods solely concentrate on static graphs, leaving the challenge of handling noise in dynamic graphs unaddressed. Our method can be regarded as the pioneering extension of graph structure learning to dynamic graph settings.

\section{Preliminaries}
This section illustrates important notations and terminology definitions in this paper.

\subsection{Notations}
We summarize the important notations in this paper and their definitions as displayed in Table \ref{notations}.
\begin{table}[h]
  \caption{Important notations in this paper.}
  \label{notations}
  \begin{tabular}{cc}
    \toprule
    Notations & Definitions  \\
    \midrule
    $\mathbf{e}_{ij}(t)$ & Feature of edge $(i, j, t)$\\
    $w_{ij}(t)$ &  Weight of edge $(i, j, t)$ \\
    \midrule
    $\mathcal{S}_{ij}(t)$  & Dynamic noise filter of edge $(i, j, t)$\\
    $h_{i}^{(l)}(t)$ & Embedding of node $i$ in the $l$-th embedding layer \\
    $\mathbf{z}_{i}(t)$ & Temporal embedding of node $i$ at time $t$ \\
    \midrule
    $\mathcal{L}_{TEL}$ & Loss function of Temporal Embedding Learner\\
    $\mathcal{L}_{DGSL}$ & Loss function of Dynamic Graph Structure Learning \\
  \bottomrule
\end{tabular}
\end{table}

\subsection{Terminology Definitions}
\begin{definition}
    \textbf{Dynamic Graph.} A dynamic graph is modeled as a sequence of timestamped events $\mathcal{G}=\{(i_1, j_1, t_1), (i_2, j_2, t_2), ...\}$, representing the addition or change of interaction between a pair of nodes at times $t_1\le t_2\le ...$ . Given node set $\mathcal{V} = \{1,2,..., |\mathcal{V}|\}$ and timestamp set $\mathcal{T} = \{t_1, t_2, ...\}$, an event between nodes $i \in \mathcal{V}$ and $j \in \mathcal{V}$ at time $t \in \mathcal{T}$ is represented by a temporal edge $(i, j, t) \in \mathcal{E}$ (edge set), whose edge feature is $\mathbf{e}_{ij}(t)$.
\end{definition}

A dynamic graph $\mathcal{G}$ also can be seen as the edge set $\mathcal{E}$ that is sorted by time order. For the remaining part of this paper, we will misuse the terminologies of ``dynamic graph $\mathcal{G}$'' and ``edge set $\mathcal{E}$'' interchangeably without distinguishing their differences. Moreover, nodes $i$ and $j$ may have multiple interactions at different times, so we consider $(i, j, t_1)$ and $(i, j, t_2)$ as different edges when $t_1 \ne t_2$.

\begin{definition}
    \textbf{Dynamic Graph Representation Learning.} Given a dynamic graph $\mathcal{G}$, dynamic graph representation learning pursues to learn a temporal mapping function $f:\mathcal{V} \times \mathcal{T} \to \mathbbm{R}^d$, where $d$ is the representation dimension and $d \ll \lvert \mathcal{V} \rvert$. The intention of the mapping function $f$ is to capture the valuable pattern of the dynamic graph for various downstream tasks.
\end{definition}

\section{Proposed method}
In this paper, edges in dynamic graphs could be noisy or disturbed. Unfortunately, mainstream solutions overlook the presence of noise in real-world dynamic graphs. This oversight is concerning because the noise in such error-prone dynamic graphs can significantly degrade their performance \cite{noiselearning2019}. To address this critical issue, we propose RDGSL, a novel representation learning method aimed at effectively absorbing and purifying noise in dynamic graphs.

Our RDGSL is mainly composed of two modules: the {\itshape Dynamic Graph Filter} that dynamically evaluates the temporal noise and generates a denoised graph, and the {\itshape Temporal Embedding Learner} that conducts representation generation that remains resilient to noise. Specifically, in the Dynamic Graph Filter, we propose the dynamic noise function, a dual-function module that dynamically evaluates and captures both the current and historical noise with the base function and temporal function, respectively. Consequently, a denoised graph is generated. Meanwhile, in the Temporal Embedding Learner, we leverage an attention mechanism that combines the evaluated temporal noise from the noise-reduced graph. It enables us to selectively focus on normal edges rather than noisy edges to obtain representation, which further increases the expressiveness of our method in downstream tasks. What's more, as illustrated in Figure \ref{framework}, we propose dynamic graph structure learning, a supervisory signal that equips our method with denoising capability. It can weaken the influence of noisy edges meanwhile strengthening the contribution of normal ones. We represent these crucial components in the following subsections.

\subsection{Dynamic Graph Filter}
In this paper, the Dynamic Graph Filter is aimed at dynamically evaluating the temporal noise and generating a denoised dynamic graph. We first introduce our dynamic noise function and then construct our Dynamic Graph Filter through an edge-weight predictor.

\subsubsection{Dynamic noise function}
\label{sec:TemporalSim}
Given that nodes with similar attributes often connect through normal edges, while noisy edges tend to link dissimilar nodes, similarity \cite{similarity2021, RSGNN2022} is a commonly used static noise function to evaluate noise in static graphs. However, noise in dynamic graphs exhibits dynamics, and using a static noise function alone cannot effectively capture the temporal aspect of noise dynamically. To tackle this challenge, we propose a dynamic noise function that incorporates both a base function and a temporal function, allowing us to address the temporal noise in dynamic graphs.

Precisely, like the static noise function, similarity acts as the base function, effectively encoding and evaluating the current noise in each incoming edge. In addition, we employ a similarity-based attention mechanism among historical interactions as the temporal function, which enables us to capture and model the historical noise, considering the evolving nature of the noise in dynamic graphs. This dual-function module empowers us to dynamically handle the temporal noise in dynamic graphs, ensuring the generation of denoised graphs even amidst the presence of noise fluctuations.

Formally, given an incoming edge, $(i,j,t)$, we define dynamic noise function $\mathcal{S}_{ij}(t)$ with the embedding of nodes $i$ and $j$ at time $t$, $\mathbf{z}_i(t)$ and $\mathbf{z}_j(t)$, which will be discribed in Section \ref{sec:TEL}:
\begin{equation}\label{temporalsimilarity} 
  \begin{aligned}
\mathcal{S}_{ij}(t) & =\underbrace{g\left(\mathbf{z}_{i}\left(t\right), \mathbf{z}_{j}\left(t\right)\right)}_{ {\text{base function}}} \\
& +\beta_{i j} \sum_{p \in \mathcal{N}^{i}(t)} \alpha_{p i}(t) g\left(\mathbf{z}_{p}\left(t^-\right), \mathbf{z}_{j}\left(t\right)\right) \kappa\left(t-t_{p}\right) \\
& +\underbrace{\left(1-\beta_{i j}\right) \sum_{q \in \mathcal{N}^{j}(t)} \alpha_{q j}(t) g\left(\mathbf{z}_{q}\left(t^-\right), \mathbf{z}_{i}\left(t\right)\right) \kappa\left(t-t_{q}\right)}_{{\text{temporal function}}},
\end{aligned}
\end{equation}
where $g(x, y) = ||x - y||^2_2$,  $p \in \mathcal{N}^{i}(t)$ and $q \in \mathcal{N}^{j}(t)$ are the historical neighbors of nodes $i$ and $j$, respectively. The term $\kappa\left(t-t_{p}\right) = \exp\left(-\delta\left(t-t_p\right)\right)$ is a time decay function where $\delta$ is a trainable parameter with the decay rate, and $t_p$ is the time point when node $i$ interact with $p$. $\alpha$ and $\beta$ are the proposed self-temporal attention and cross-temporal attention respectively, and we will describe them as follows.

\textbf{Self-temporal attention.}
The historical noise experienced by nodes themselves can exert varying effects on their current representations. For example, if a node has encountered noisy interactions in the past, it can contaminate its current representation. Therefore, we introduce the self-temporal attention mechanism, enabling our method to capture the historical noise patterns that persist within the nodes themselves. The term $g(p,j)=||\mathbf{z}_p(t^-)-\mathbf{z}_j(t)||_2^2$ represents the similarity between $i$'s certain historical behavior $(i,p,t_p)$ and the current behavior $(i,j,t)$, describing historical noise influence on current event. We define self-temporal attention as follows:
\begin{equation}\label{selftemporal1}
  \tilde{\alpha}_{p i}(t)=\sigma\left(\kappa\left(t-t_{p}\right) \mathbf{a}^{\top}\left[\mathbf{W} \mathbf{z}_{i}\left(t\right) || \mathbf{W} \mathbf{z}_{p}\left(t\right)\right]\right),
\end{equation}
\begin{equation}\label{selftemporal2}
  \alpha_{p i}(t)=\frac{\exp \left(\tilde{\alpha}_{p i}(t)\right)}{\sum_{p^{\prime} \in \mathcal{N}^{i}(t)} \exp \left(\tilde{\alpha}_{p^{\prime} i}(t)\right)},
\end{equation}
where $\sigma(\cdot)$ is the sigmoid function. $\kappa\left(t-t_{p}\right)$ is the time decay function where $p$ will have a larger impact on event occurred at $t$ if $t_p$ is closer to $t$. $\mathbf{a}$ and $\mathbf{W}$ are learnable parameters, and $||$ is concatenation operation. Similarly, we can get the expression of~$\alpha_{q j}(t)$.

\textbf{Cross-temporal attention.} 
The historical noise experienced by the historical neighbors of a node can also have diverse effects on its current representation. When the historical neighbors of a node have engaged in noisy interactions in the past, it may pollute the representation of this certain node. Hence, we propose cross-temporal attention to capture the historical noise patterns in historical neighbors. Firstly, the information of $i$'s neighbors is:
\begin{equation}\label{crosstemporal1}
  \tilde{\mathbf{z}}_{i}\left(t\right)=\sigma\left(\sum_{p \in \mathcal{N}^{i}(t)} \alpha_{p i}(t) \mathbf{W} \mathbf{z}_{i}\left(t\right)\right).
\end{equation}

Then, the average of the time decay is calculated as $\overline{\delta {t_{p}}}=\frac{1}{\left|\mathcal{N}^{i}(t)\right|} \sum_{p \in \mathcal{N}^{i}(t)}\left(t-t_{p}\right)$, and the final cross-temporal attention represents as follows:
\begin{equation}\label{crosstemporal2}
  \tilde{\beta}_{i}=s\left(\kappa\left(\overline{\delta{t_{p}}}\right) \tilde{\mathbf{z}}_{i}\left(t\right)\right), \quad \tilde{\beta}_{j}=s\left(\kappa\left(\overline{\delta{t_{p}}}\right) \tilde{\mathbf{z}}_{j}\left(t\right)\right),
\end{equation}
\begin{equation}\label{crosstemporal3}
  \beta_{i j}=\frac{\exp (\tilde{\beta}_{i})}{\exp (\tilde{\beta}_{i})+\exp (\tilde{\beta}_{j})},
\end{equation}
where $s\left( \cdot\right)$ is a neural network.

\subsubsection{Building Dynamic Graph Filter}
\label{sec:DGC}
To generate a denoised graph, we assign an edge-weight predictor for each incoming edge. Given a dynamic graph $\mathcal{G}$, our denoised graph is weighted, and the weight of a temporal edge $(i, j, t)$ is represented as $w_{ij}\left(t\right) \in \mathbbm{R}^+$. Moreover, noisy edges on denoised graphs are assigned low weights while clean edges are assigned high weights. In practice, we compute the edge weight by a multi-layer procedure (MLP) between nodes $i$ and $j$ at time $t$ with their representations, $\mathbf{z}_i(t)$ and $\mathbf{z}_j(t)$:
\begin{equation}
\label{weight}  w_{ij}\left(t\right)=\operatorname{ReLU}\left(\operatorname{MLP}\left(\mathbf{z}_i\left(t\right) \| \mathbf{z}_j\left(t\right)\right)\right),
\end{equation}
where $\|$ is the concatenation operator. The representation of node $i$ at time $t$, $\mathbf{z}_i(t)$, is learned from the Temporal Embedding Learner in Section \ref{sec:TEL} and initialized by the zero vector. For simplicity, we utilize an MLP as our Dynamic Graph Filter, which benefits from the training of dynamic graph structure learning and effectively generates a denoised graph, which will be described in Section~\ref{sec:loss}.

\subsection{Temporal Embedding Learner}
\label{sec:TEL}
To prevent the issue of more severe noise mentioned before, we deploy the Temporal Embedding Learner to generate representations that are robust to noise for downstream tasks. The critical challenge for this module lies in how to effectively utilize denoised graphs to enhance the quality of representation generation. To address this challenge, we integrate an attention mechanism with the evaluated temporal noise on the noise-reduced graphs. It can obtain representation by selectively focusing on aggregating the neighbors' information from normal edges instead of noisy edges, bolstering the effectiveness and expressiveness in handling noisy dynamic graphs. This integration ensures that our representations are resilient to noise and well-suited for downstream tasks.

Different from traditional TGN, for each iteration, the memory module is initialized by the temporal node representation, \textit{i.e.}, $\mathbf{z}_i(t)$ and $\mathbf{z}_j(t)$, and the weight of the edge between them, \textit{i.e.}, $w_{ij}(t)$. Then, we combine an $L$-layer temporal graph attention network \cite{TGN2020} with the edge weight we evaluated from Dynamic Graph Filter to aggregate neighborhood information as:
\begin{equation}\label{TEL1}
  \mathbf{h}_{i}^{(l)}(t)=\mathbf{MLP}_{2}^{(l)}\left(\mathbf{h}_{i}^{(l-1)}(t) \| \tilde{\mathbf{h}}_{i}^{(l)}(t)\right),
\end{equation}
\begin{equation}\label{TEL2}
  \tilde{\mathbf{h}}_{i}^{(l)}(t)=ReLU\left(\sum_{j \in \mathcal{N}^i} w_{ij}(t) \mathbf{MLP}_{1}^{(l)}\left(\mathbf{h}_{j}^{(l-1)}(t)\left\|\mathbf{e}_{i j}\right\| \phi\left(t-t_{j}\right)\right)\right),
\end{equation}
where $\mathbf{MLP}_1^{(l)}, \mathbf{MLP}_2^{(l)}$ are two different MLPs, $\mathbf{e}_{ij}(t)$ is the edge feature and $\phi(\cdot)$ is time encoding presented in \cite{time2vec2019} and used in \cite{TGAT2020}. Temporal node representation is recorded as $\mathbf{z}_i(t)=\mathbf{h}_i^{(L)}(t)$, and $L$ is the number of attention layer.

Existing TGNs utilize link prediction as the self-supervised task for temporal representation generation. Since we assume edges are noisy and disturbed in this paper, evolving node classification is conducted to effectively reduce the impact of noise. This measure ensures that our method is better equipped to handle noisy dynamic graphs. Thus, the training loss of Temporal Embedding Learner is recorded as $\mathcal{L}_{TEL}$:
\begin{equation}\label{lossTEL}
  \mathcal{L}_{TEL}=\sum_{i \in \mathcal{V}}{\mathbf{CE}\left(\hat{y}_i(t), y_i(t)\right)},
\end{equation}
where $\mathbf{CE}(\cdot)$ is the cross-entropy. $\hat{y}_i(t)=\operatorname{MLP}(\mathbf{z}_i(t))$ is the predicting evolving label of node $i$, and $y_i(t)$ is the ground truth.

\subsection{Dynamic Graph Structure Learning}
\label{sec:loss}
We propose dynamic graph structure learning to ensure that our method has the ability to denoise effectively in dynamic graphs. As previously mentioned, we build Dynamic Graph Filter and Temporal Embedding Learner in our RDGSL. The challenge of dynamic graph structure learning is how to conduct a supervisory signal to optimize these two components that equip them has the ability to attenuate the influence of noisy edges while enhancing the contribution of normal edges.

To address this challenge, we consider the edge weight $w_{ij}(t)$ for a given temporal edge $(i, j, t)$ in dynamic graphs. Specifically, we modify the edge weight to weaken the impact of noisy edges by reducing their values while increasing the edge weight of normal edges to strengthen their contribution. However, directly using an edge (an event) itself only focuses on the event participants, which neglects the importance of non-participants. To address this problem meanwhile controlling the computational cost, negative sampling \cite{negativeSampling2013} is introduced: For each incoming edge $(i, j, t)$, $Q$ randomly sampled nodes, $\{n_1, n_2, ..., n_Q\} \subset \mathcal{V}$, and equally random-sampled time points, $\{t_{n_1}, t_{n_2}, ..., t_{n_Q}\} \subset \mathcal{T}$, are used as negative samples to construct the negative events, $\{(i, n_1, t_{n_1}), ..., (i, n_Q, t_{n_Q}))\}$, which not happened actually.

Therefore, the positive and negative events are re-weighted by our method with the supervisory signal based on structure learning. For a positive sample $j$ of node $i$, $j \in \mathcal{N}^i$, $-\log \sigma\left(\frac{\mathcal{S}_{ij}\left(t\right)}{\epsilon}\right)(w_{ij}(t)-1)^{2}$ will be minimized where $\mathcal{S}_{ij}(t)$ is the dynamic noise function mentioned in Section \ref{sec:TemporalSim} and $\epsilon \in \mathbbm{R}^+$ is a hyper-parameter to control it, and $\sigma(\cdot)$ is the sigmoid function. If two nodes are similar, they are more likely to share a clean edge and $-\log \sigma\left(\frac{\mathcal{S}_{ij}\left(t\right)}{\epsilon}\right)$ would be large, which will force $w_{ij}(t)$ close to one by minimizing this signal. Instead, if two nodes are dissimilar, they tend to connect with a noisy edge and $-\log \sigma\left(\frac{\mathcal{S}_{ij}\left(t\right)}{\epsilon}\right)$ would be small, which will have little influence on $w_{ij}(t)$ with the signal minimization. Similarly, for a negative sample $n_q$ of node $i$, $-\log \sigma\left(\frac{-\mathcal{S}_{in_q}\left(t_{n_q}\right)}{\epsilon}\right)(w_{in_q}(t_{n_q})-0)^{2}$ will be minimized. If two nodes are dissimilar, $-\log \sigma\left(\frac{-\mathcal{S}_{in_q}\left(t_{n_q}\right)}{\epsilon}\right)$ would be large, which will force $w_{in_q}(t_{n_q})$ to zero by minimizing this signal as expected. In summary, dynamic graph structure learning is proposed as $\mathcal{L}_{DGSL}$:
\begin{equation}\label{lossDGC}
  \begin{aligned}
\mathcal{L}_{DGSL}=\sum_{i \in \mathcal{V}} & \sum_{j \in \mathcal{N}^i\left(t\right)}\left[-\log \sigma\left(\frac{\mathcal{S}_{ij}\left(t\right)}{\epsilon}\right)(w_{ij}(t)-1)^{2}\right. \\
& \left.-\sum_{q=1}^{Q} \cdot \mathbb{E}_{n_q \sim P_{n}\left(i\right)} \log \sigma\left(\frac{-\mathcal{S}_{in_q}\left(t_{n_q}\right)}{\epsilon}\right)(w_{in_q}(t_{n_q})-0)^{2}\right],
\end{aligned}
\end{equation}
where $n_q \sim P_{n}\left(i\right)$ is the distribution of negative samples of node $i$.

The final training objective of RDGSL is defined as:
\begin{equation}\label{lossFinal}
  \underset{\theta_{S}, \theta_{\mathcal{G}}}{\arg \min } \mathcal{L}_{TEL}+\gamma \mathcal{L}_{DGSL},
\end{equation}
where $\theta_{S}, \theta_\mathcal{G}$ are parameters of Dynamic Graph Filter and Temporal Embedding Learner, respectively, and $\gamma \in \mathbbm{R}^+$ is a hyper-parameter to balance the supervisory signal from dynamic graph structure learning and the evolving node classification task. The method proposed in this paper is end-to-end, which can resist the temporal noise in dynamic graphs with structure learning.

\subsection{Complexity Analysis}
We analyze the time complexity of training RDGSL with respect to the input data. Denote the number of nodes as $\sharp\mathcal{V}$, the edges as $\sharp\mathcal{E}$, and the dimension of the input features as $d$. Similar to TGNs \cite{TGN2020}, we utilize the memory module to update nodes with edges and classify them in a recursive manner. The time complexity of the memory module is $\mathcal{O}\left(d \cdot \sharp \mathcal{E}\right)$, which is also the time complexity of TGNs. If we achieve Dynamic Graph Filter by feeding the input data from the beginning time to evaluate the temporal noise and denoise, the corresponding time complexity is:
\begin{equation}
    \label{complexity1}
    \mathcal{O}\left(\sum_{e=0}^{\sharp \mathcal{E}-1} d e\right)=\mathcal{O}\left(d\left((\sharp \mathcal{E})^{2}-\sharp \mathcal{E}\right) / 2\right)=\mathcal{O}\left(d(\sharp \mathcal{E})^{2}\right),
\end{equation}
which is much higher than the complexity of TGNs.

Therefore, in Dynamic Graph Filter, we record the weight of edges that have appeared (the memory complexity is $\mathcal{O}\left(\sharp\mathcal{E}\right)$).  In every batch $b$, the time complexity is $\mathcal{O}\left(d \cdot \sharp\mathcal{E}^{(b)}\right)$, where $\mathcal{E}^{(b)}$ is the edges set of batch $b$ to be denoised. Summing the complexity of all batches up leads to $\sum_{b} d \cdot \sharp \mathcal{E}^{(b)}$. Thus, the final time complexity of RDGSL is
\begin{equation}\label{comlexity2}
\begin{aligned}
    \mathcal{O}\left(\mathbb{E}\left[d \cdot \sharp \mathcal{E} + \sum_{b} d \cdot \sharp \mathcal{E}^{(b)}\right]\right)& =\mathcal{O}\left(d \cdot \sharp \mathcal{E}+ d \cdot \mathbb{E}\left[\sum_{b}\sharp \mathcal{E}^{(b)}\right]\right) \\
    & =\mathcal{O}(2 d \cdot \sharp \mathcal{E})=\mathcal{O}(d \cdot \sharp \mathcal{E}),
\end{aligned}
\end{equation}
which is the same as TGNs.

\begin{table}
  \caption{Details of datasets.}
  \label{datasets}
  \begin{tabular}{ccccc}
    \toprule
    Datasets & \#Nodes & \#Edges & \#Edge feature & Label type\\
    \midrule
    Wikipedia & 9,227 & 157,474 & 172 &editing ban\\
    Reddit & 10,984 & 672,447 & 172 &posting ban\\
    MOOC & 7,144 & 411,749 &100 &course dropout\\
  \bottomrule
\end{tabular}
\end{table}

\begin{table*}
  \caption{\centering Evolving node classification performance (AUC(\%) $\pm$ Std) on various types of noisy dynamic graphs with both static graph methods (above) and dynamic graph methods (below). * denotes the methods that have the ability to denoise. Note that we abridge the name of perturbation methods due to space limitations.}
  \label{evolvingClassification}
  \small
  \setlength{\tabcolsep}{1.0mm}{
  \begin{tabular}{ccccc|cccc|ccccc}
    \toprule
    \multirow{2}{*}{\textbf{Models}}
     & \multicolumn{4}{c}{ \textbf{Wikipedia}} & \multicolumn{4}{c}{\textbf{Reddit}} & \multicolumn{4}{c}{\textbf{MOOC}} \\
     & \textbf{Original} & \textbf{Time} & \textbf{Feature} & \textbf{Structure} & \textbf{Original} & \textbf{Time} & \textbf{Feature} & \textbf{Structure} & \textbf{Original} & \textbf{Time} & \textbf{Feature} & \textbf{Structure} \\
    \midrule
    \textbf{GraphSage} & 82.14$\pm$0.2 & 69.88$\pm$1.0 & 73.28$\pm$0.6 & 68.23$\pm$0.1 & 61.62$\pm$0.1 & 55.71$\pm$0.3 & 56.36$\pm$0.9 & 56.01$\pm$0.5 & 62.88$\pm$0.1 & 56.56$\pm$0.7 & 55.19$\pm$1.0 & 54.38$\pm$0.6 \\
    \textbf{Pro-GCN*}  & 82.37$\pm$0.1 & 67.88$\pm$0.2 & 72.89$\pm$0.1 & 67.77$\pm$0.1 & 62.28$\pm$0.1 & 57.95$\pm$2.4 & 55.71$\pm$0.4 & 56.29$\pm$1.0 & 60.95$\pm$1.0 & 53.98$\pm$1.0 & 56.10$\pm$0.1 & 54.55$\pm$0.8 \\
    \textbf{SimP-GCN*} & 82.48$\pm$0.4 & 74.87$\pm$1.0 & 79.25$\pm$0.2 & 79.21$\pm$0.1 & 63.03$\pm$0.1 & 58.24$\pm$0.1 & 57.37$\pm$0.2 & 54.60$\pm$0.4 & 63.99$\pm$0.6 & 57.61$\pm$0.2 & 57.20$\pm$0.6 & 55.91$\pm$1.0 \\
    \textbf{RSGNN*}    & 83.14$\pm$0.1 & 77.72$\pm$0.6 & 80.04$\pm$0.5 & 75.77$\pm$0.6 & 63.82$\pm$0.1 & 60.38$\pm$0.2 & 58.98$\pm$0.5 & 56.60$\pm$0.5 & 64.49$\pm$0.1 & 59.18$\pm$0.1 & 61.86$\pm$0.1 & 58.33$\pm$0.1 \\
    \midrule
    \textbf{CTDNE} & 75.89$\pm$0.5 & 66.20$\pm$0.7 & 69.31$\pm$0.6 & 60.78$\pm$0.5 & 59.43$\pm$0.6 & 51.95$\pm$0.7 & 54.84$\pm$0.6 & 50.29$\pm$0.9 & 67.54$\pm$0.7 & 56.60$\pm$0.8 & 56.99$\pm$0.8 & 53.89$\pm$0.6\\
    \textbf{DyRep} & 84.59$\pm$2.2 & 79.96$\pm$2.5 & 76.75$\pm$2.1 & 76.08$\pm$2.0 & 62.91$\pm$2.4 & 56.66$\pm$1.9 & 54.74$\pm$2.5 & 50.62$\pm$2.8 & 67.76$\pm$0.5 & 57.99$\pm$1.4 & 62.62$\pm$1.2 & 60.27$\pm$1.2\\
    \textbf{JODIE} & 84.84$\pm$1.2 & \underline{80.66$\pm$0.8} & 79.01$\pm$0.9 & 78.49$\pm$1.4 & 61.83$\pm$2.7 & 52.79$\pm$2.4 & 53.84$\pm$2.5 & 49.20$\pm$2.5 & 66.87$\pm$0.4 & 61.83$\pm$1.5 & 62.19$\pm$1.5 & 60.11$\pm$1.7\\
    \textbf{TGAT}  & 83.69$\pm$0.7 & 74.94$\pm$0.9 & 74.60$\pm$0.9 & 72.03$\pm$0.5 & 65.56$\pm$0.7 & 60.61$\pm$0.9 & 60.08$\pm$0.5 & \underline{59.80$\pm$0.8} & 53.95$\pm$0.2 & 52.82$\pm$0.7 & 51.48$\pm$0.9 & 51.38$\pm$0.8\\
    \textbf{TGN}   & \underline{87.81$\pm$0.3} & 79.50$\pm$0.4 & \underline{82.99$\pm$0.7} & \underline{78.77$\pm$0.5} & 67.06$\pm$0.9 & \underline{62.63$\pm$0.6} & \underline{60.30$\pm$0.9} & 57.56$\pm$0.8 & \underline{69.54$\pm$1.0} & \underline{64.82$\pm$0.8} & \underline{64.52$\pm$0.9} & \underline{61.24$\pm$0.7}\\
    \textbf{PINT}  & 87.59$\pm$0.6 & 77.86$\pm$0.3 & 81.88$\pm$0.4 & 75.56$\pm$0.8 & \underline{67.31$\pm$0.2} & 62.25$\pm$0.3 & 58.35$\pm$1.0 & 55.26$\pm$0.8 & 68.77$\pm$1.1 & 62.72$\pm$1.2 & 63.81$\pm$0.8 & 59.11$\pm$0.2 \\
    \textbf{Ours*}  & \textbf{89.85$\pm$0.3} & \textbf{86.98$\pm$0.2} & \textbf{87.33$\pm$0.1} & \textbf{86.10$\pm$0.4} & \textbf{68.79$\pm$0.8} & \textbf{65.98$\pm$0.6} &\textbf{65.43$\pm$0.9} & \textbf{64.12$\pm$0.9} & \textbf{72.03$\pm$0.9} & \textbf{71.10$\pm$0.9} & \textbf{71.87$\pm$0.9} & \textbf{70.94$\pm$0.9}\\
    \bottomrule
  \end{tabular}}
\end{table*}

\begin{table*}
  \caption{\centering Temporal link prediction performance (Accuracy(\%) $\pm$ Std) on various types of noisy dynamic graphs with dynamic graph methods. }
  \label{linkPrediction}
  \small
  \begin{tabular}{ccccccccc}
    \toprule
    \textbf{Datasets} & 
    \textbf{Graph} & \textbf{CTDNE} & \textbf{DyRep} & \textbf{JODIE} & \textbf{TGAT} & \textbf{TGN} & \textbf{PINT} & \textbf{Ours}\\
    \midrule
    \multirow{4}{*}{\textbf{Wikipedia}}
    & \textbf{Original graph} & 79.42 $\pm$ 0.4 & 87.77 $\pm$ 0.2 & 87.04 $\pm$ 0.4 & 88.14 $\pm$ 0.2 & 89.51 $\pm$ 0.4 & 89.95 $\pm$ 0.1 & 89.25 $\pm$ 0.4 \\
    & \textbf{Disturb time} & 69.61 $\pm$ 0.5 & 79.22 $\pm$ 0.4 & 80.90 $\pm$ 0.6 & 76.98 $\pm$ 0.3 & \underline{81.36 $\pm$ 0.5} & 76.88 $\pm$ 0.4 & \textbf{86.90 $\pm$ 0.7} \\
    & \textbf{Disturb edge feature} & 73.07 $\pm$ 0.2 & 77.89 $\pm$ 0.3 & \underline{81.91 $\pm$ 0.6} & 78.72 $\pm$ 0.2 & 81.64 $\pm$ 0.2 & 78.52 $\pm$ 0.8 & \textbf{86.51 $\pm$ 0.6} \\
    & \textbf{Disturb structure} & 67.88 $\pm$ 0.1 & 76.49 $\pm$ 0.4 & 79.19 $\pm$ 0.4 & 76.10 $\pm$ 0.2 & \underline{80.09 $\pm$ 0.7} & 76.42 $\pm$ 0.1 & \textbf{85.10 $\pm$ 0.7}\\
    \midrule
    \multirow{4}{*}{\textbf{Reddit}}
    & \textbf{Original graph} & 73.76 $\pm$ 0.5 & 92.11 $\pm$ 0.2 & 90.91 $\pm$ 0.3 & \underline{92.92 $\pm$ 0.3} & 92.56 $\pm$ 0.2 & 91.39 $\pm$ 0.1 & \textbf{93.28 $\pm$ 0.3} \\
    & \textbf{Disturb time} & 68.92 $\pm$ 0.6 & 83.11 $\pm$ 0.3 & 81.49 $\pm$ 0.4 & 79.33 $\pm$ 0.3 & \underline{83.21 $\pm$ 0.3} & 81.40 $\pm$ 0.2 & \textbf{91.58 $\pm$ 0.4} \\
    & \textbf{Disturb edge feature} & 69.61 $\pm$ 0.5 & 82.95 $\pm$ 0.3 & \underline{83.17 $\pm$ 0.3} & 81.87 $\pm$ 0.3 & 82.07 $\pm$ 0.5 & 79.86 $\pm$ 0.2 & \textbf{91.22 $\pm$ 0.2} \\
    & \textbf{Disturb structure} & 68.22 $\pm$ 1.0 & 79.79 $\pm$ 0.1 & 83.84 $\pm$ 0.2 & 78.63 $\pm$ 0.4 & \underline{83.89 $\pm$ 0.1} & 82.38 $\pm$ 0.6 & \textbf{90.07 $\pm$ 0.6} \\
    \midrule
    \multirow{4}{*}{\textbf{MOOC}}
    & \textbf{Original graph} & 65.34 $\pm$ 0.7 & 73.36 $\pm$ 0.4 & 76.45 $\pm$ 0.6 & 75.20 $\pm$ 0.5 & 81.83 $\pm$ 0.6 & 80.98 $\pm$ 0.1 & 81.62 $\pm$ 0.2 \\
    & \textbf{Disturb time} & 59.42 $\pm$ 0.8 & 66.91 $\pm$ 0.5 & 67.20 $\pm$ 1.0 & 71.51 $\pm$ 0.7 & \underline{75.91 $\pm$ 0.7} & 70.65 $\pm$ 0.2 & \textbf{79.76 $\pm$ 0.5} \\
    & \textbf{Disturb edge feature} & 57.81 $\pm$ 0.9 & 64.28 $\pm$ 0.4 & 69.01 $\pm$ 0.5 & 65.11 $\pm$ 0.4 & \underline{75.48 $\pm$ 0.6} & 75.28 $\pm$ 0.2 & \textbf{80.21 $\pm$ 0.4} \\
    & \textbf{Disturb structure} & 56.08 $\pm$ 1.3 & 64.81 $\pm$ 0.5 & 66.82 $\pm$ 0.6 & 62.89 $\pm$ 0.4 & \underline{73.18 $\pm$ 0.7} & 72.39 $\pm$ 0.1 & \textbf{79.97 $\pm$ 0.5}\\
    \bottomrule
  \end{tabular}
\end{table*}

\section{Experiments}
\subsection{Experimental Settings}
\label{sec:ExpSetting}
\subsubsection{Datasets.} In this paper, three widely-used real-world dynamic graph datasets are employed, including Wikipedia \cite{JODIE2019}, Reddit \cite{reddit2020} and MOOC \cite{JODIE2019}. Note that the train-validation-test split on all datasets is the same as \cite{TGN2020}, which is convenient for us to analyze and compare. Data between the train split and validation or test split has no intersection. The detailed descriptions of our datasets are shown in Table \ref{datasets}.

\subsubsection{Perturbation Methods.}\label{sec:noisyGraph} For verifying the robustness of our method under different types of noise, we put forward four perturbation methods on dynamic graphs. \textbf{Original graph:} The dynamic graph of original datasets, which may exist inherent noise naturally. \textbf{Disturb time:} Edges are picked randomly with perturbation rate $p$, whose timestamps are perturbed by Gaussian noise \cite{MATA2021}. \textbf{Disturb edge feature:} We pick edges with perturbation rate $p$. For every selected edge, each dimension of their features may be forced to zero value with probability $p$. \textbf{Disturb structure:} With the rate $p$, original edges are omitted, and fake edges are added randomly in dynamic graphs. The perturbation method used in our experiments is widely used in static graphs denoising field \cite{RSGNN2022, similarity2021}, which simulates the real-world scenario and also provides a controlled environment. Our perturbation method can evaluate and compare the ability of different methods in handling noisy dynamic graphs.

\subsubsection{Baselines.} Our baselines are from two fields: static graph methods and dynamic graph methods. The static graph methods aim to evaluate the denoising ability of our method, including GraphSage \cite{GraphSage2017}, Pro-GCN \cite{GSL_0_2020}, SimP-GCN \cite{SimP-GCN2020}, RSGNN \cite{RSGNN2022}. Specifically, GraphSage is a highly effective static graph method that is often used as a baseline to compare with dynamic graph methods, and Pro-GCN is a classic denoising method with structure learning in static graphs. RSGNN and SimP-GCN are the state-of-the-art static structure learning methods used for denoising, both of which utilize similarity as the static noise function. To adapt static graph methods to dynamic graph settings, in practice, for one epoch, we divide the timestamps in dynamic graphs into several time intervals. We then record the events and labels that appeared before the end of each interval, which enables us to construct static graphs for conducting node classification tasks using these static graph methods. Afterward, we utilize the LSTM \cite{LSTM1997} to update the node representations as the initial node features for the next intervals. Repeat these steps until the end of the data. On the other hand, state-of-the-art dynamic graph methods in representation learning are chosen as follows: CTDNE \cite{CTDNE2018}, DyRep \cite{DyRep2019}, JODIE \cite{JODIE2019}, TGAT \cite{TGAT2020}, TGN \cite{TGN2020}, PINT \cite{pint2022}. The detailed discrimination of dynamic graph methods is omitted due to space limitations. It is worth noting that all of the static graph methods are primarily designed for node classification tasks, thus, in our evaluation, we do not consider them in the evaluation of the temporal link prediction task.

\begin{figure*}
  \centering
  \begin{subfigure}{0.49\linewidth}
    \centering
    \includegraphics[width=\linewidth]{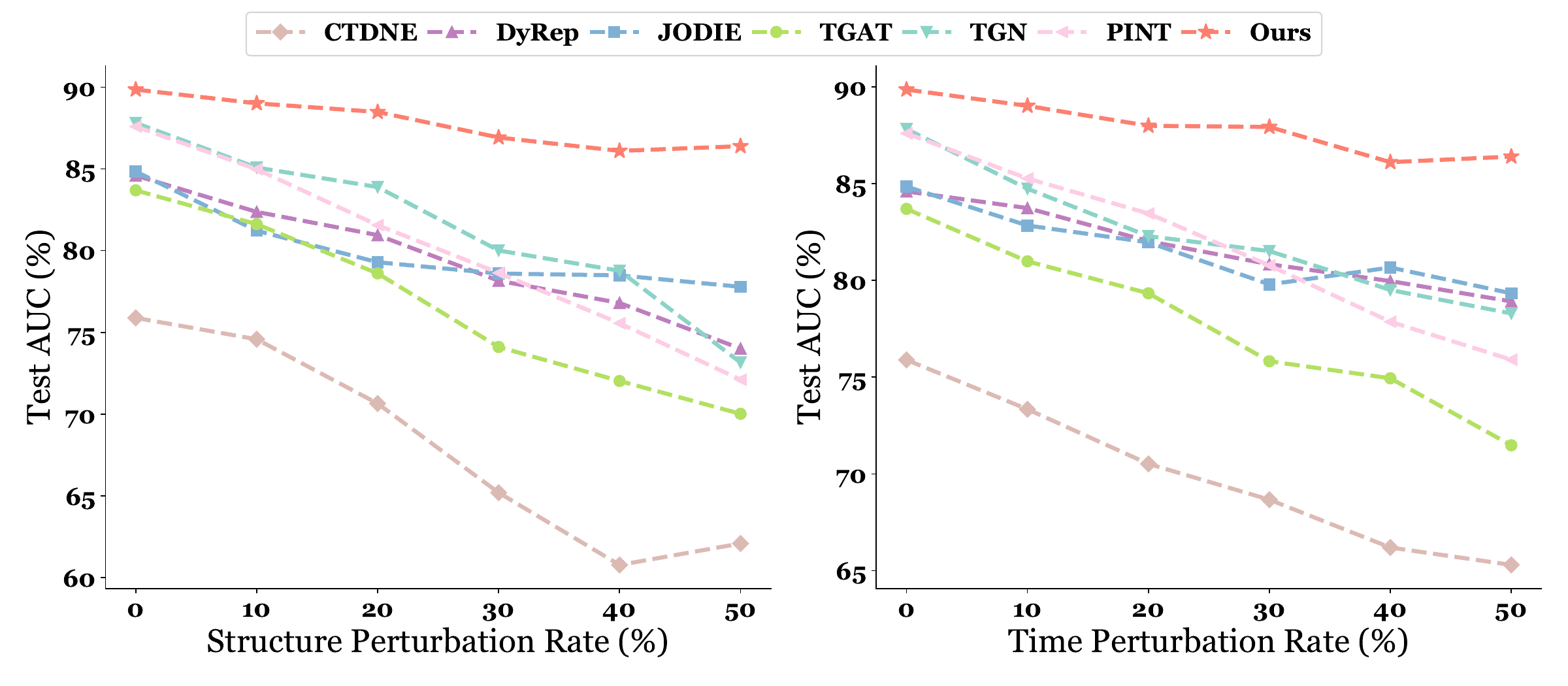}
    \caption{Wikipedia.}
    \label{time_ptb}
  \end{subfigure}
  \centering
  \begin{subfigure}{0.49\linewidth}
    \centering
    \includegraphics[width=\linewidth]{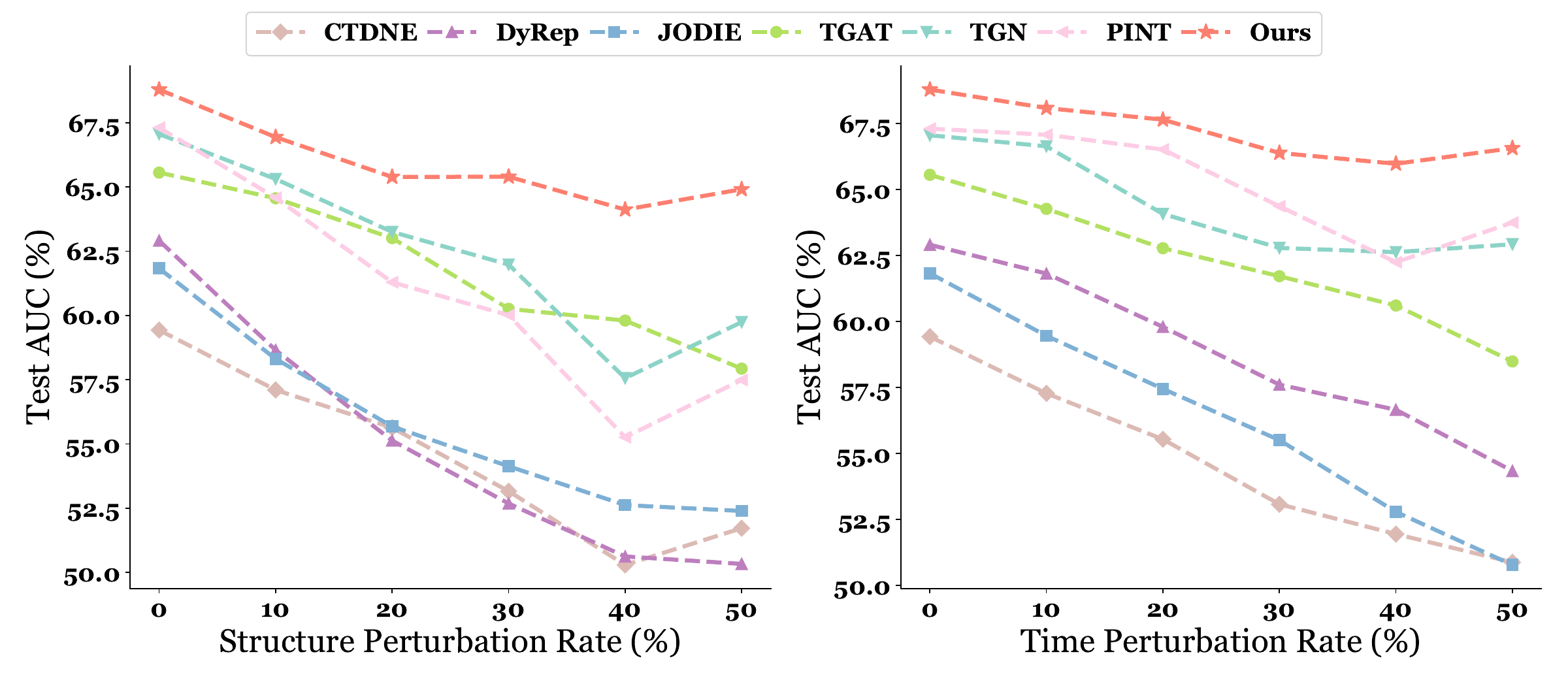}
    \caption{Reddit.}
    \label{structure_ptb}
  \end{subfigure}
  \caption{Robustness under different levels of perturbation rates with structure perturbation and time perturbation.}
  \label{different_ptb_rate}
\end{figure*}

\subsection{Evolving Classification on Noisy Graphs}
\label{sec:classification}
We begin our study by employing four perturbation methods (mentioned in Section \ref{sec:noisyGraph}) on each of our three dynamic graphs. The perturbation probability $p$ for all methods is set to 0.4. As a result, we generate a total of 12 dynamic graphs, each containing noise. We then proceed to conduct the evolving node classification task on these noisy dynamic graphs and report the results in terms of AUC ROC. Importantly, we maintain consistent experimental settings with TGN \cite{TGN2020}, and for all baselines on the original graphs, we refer to their corresponding table numbers as reported in \cite{TGN2020}.

Our experimental results are shown in Table \ref{evolvingClassification}. Notably, RDGSL outperforms all other baselines on both the original graphs and the perturbation graphs, highlighting the remarkable effectiveness of RDGSL. Moreover, when noise is introduced, the impact of noise on RDGSL's performance is negligible, whereas the baselines exhibit a considerable decline, underscoring RDGSL's successful noise absorption capabilities. Among the baselines, most static graph methods, despite having denoising capabilities, have suboptimal results compared to dynamic graph methods. This may be because static graph methods are unable to effectively capture the temporal noise in dynamic graphs, leading to lower performance.

\subsection{Link Prediction on Noisy Graphs}
\label{sec:linkprediction}
In addition to evolving classification, the temporal representations we obtained can also be effectively utilized in the link prediction task. For this task, we consider the current edges as positive samples and randomly sample an equal number of negative edges that do not exist. Subsequently, we train a two-layer MLP to classify these positive and negative edges, yielding the probability of whether the edge occurs or not, as done in TGN \cite{TGN2020}. 

The accuracy results are presented in Table \ref{linkPrediction}. In the case of original graphs, RDGSL achieves the highest accuracy among all baselines on Reddit and also remains highly competitive on Wikipedia and MOOC. It may be owing to the fact that the inherent noise in Reddit is strong, leading to poor performance in baselines but RDGSL. Notably, for perturbation graphs, RDGSL consistently outperforms all baselines by a significant margin, which demonstrates that RDGSL achieves more satisfactory results under more devastating noise.

\subsection{Robust of Different Perturbation Rates}
We design a comprehensive set of experiments to assess the performance of RDGSL under varying levels of perturbation rates. Specifically, we explore perturbation rates ranging from 0 to 0.5 with an interval of 0.1. Due to the consistent findings across datasets, perturbation methods, and downstream tasks, we only present the test AUC results for evolving node classification conducted on Wikipedia and Reddit with structure and time perturbation.

The results are visually depicted in Figure \ref{different_ptb_rate}. Notably, RDGSL consistently outperforms all other baselines across all perturbation rates. As the perturbation rate increases, the effectiveness of all baselines significantly declines. Conversely, RDGSL exhibits remarkable stability even under higher perturbation rates, showcasing its robustness and providing further evidence of RDGSL's capacity to effectively denoise on noisy dynamic graphs.

\begin{figure}
  \centering
  \begin{subfigure}{0.49\linewidth}
    \centering
    \includegraphics[width=\linewidth]{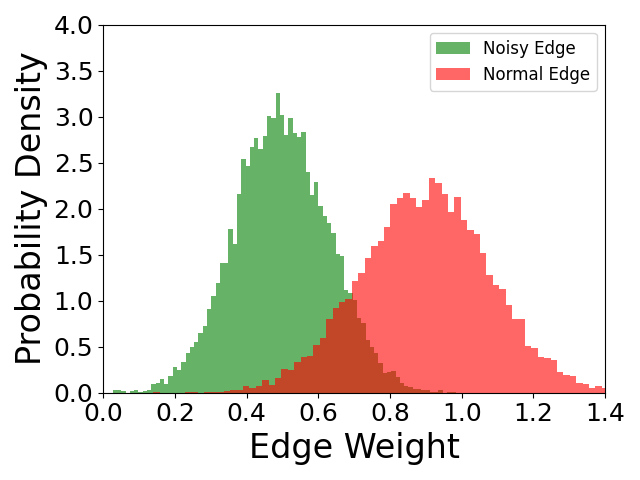}
    \caption{Wikipedia.}
    \label{wiki}
  \end{subfigure}
  \centering
  \begin{subfigure}{0.49\linewidth}
    \centering
    \includegraphics[width=\linewidth]{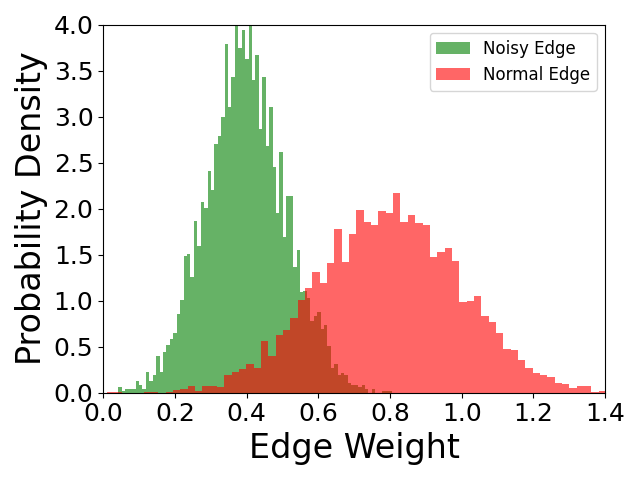}
    \caption{MOOC.}
    \label{mooc}
  \end{subfigure}
  \caption{Distribution of edge weights on noisy/normal edges.}
  \label{edge_weight}
\end{figure}

\subsection{Analysis of the Denoised Graphs}
To provide insights into how RDGSL effectively distinguishes and assigns weights to edges in dynamic graphs, we conduct an analysis of the edge weight distribution in the denoised graphs. Specifically, we extract the weights of edges in the denoised graphs obtained from the best epoch of the Dynamic Graph Filter. Subsequently, we plot the distribution of edge weights of normal edges (the edges we do not disturb) and noisy edges separately. Note that our method itself does not have prior knowledge about whether a given edge is normal or noisy.

The results obtained on Wikipedia and MOOC are displayed in Figure \ref{edge_weight}. We observe that the weights assigned to normal edges are notably larger than those to noisy edges. It suggests that the influence of normal edges is significantly greater, highlighting RDGSL's remarkable capacity to effectively absorb and purify noise in dynamic graphs. Moreover, it is interesting to note that even in the case of normal edges, there still exist many edges with relatively low weights. It may be attributed to the fact that normal edges that we do not disturb may also contain inherent noise, prompting RDGSL to assign them lower weights to mitigate their potential adverse effects.

\subsection{Ablation Study}
\label{sec:ablationstudy}
We conduct ablation experiments by analyzing the contribution of various components in RDGSL. In detail, we conduct four variants including the RDGSL-DGF, the RDGSL-TEL, the RDGSL-Sim, and the w/o DGSL. The RDGSL-DGF only contains the Dynamic Graph Filter module to verify its impact on our method, while the RDGSL-TEL preserves the Temporal Embedding Learner to test its contribution to the final results. Furthermore, we also implement RDGSL-Sim where we replace the dynamic noise function with the traditional GSL-based static noise function \cite{RSGNN2022, similarity2021}. With the comparison between the RDGSL and the RDGSL-Sim, the advantage of our dynamic noise function could be presented. Moreover, we set $\gamma = 0$ in Equation \ref{lossFinal} to perform the w/o DGSL, examining the effectiveness of dynamic graph structure learning. 

\begin{figure}
  \centering
  \includegraphics[width=\linewidth]{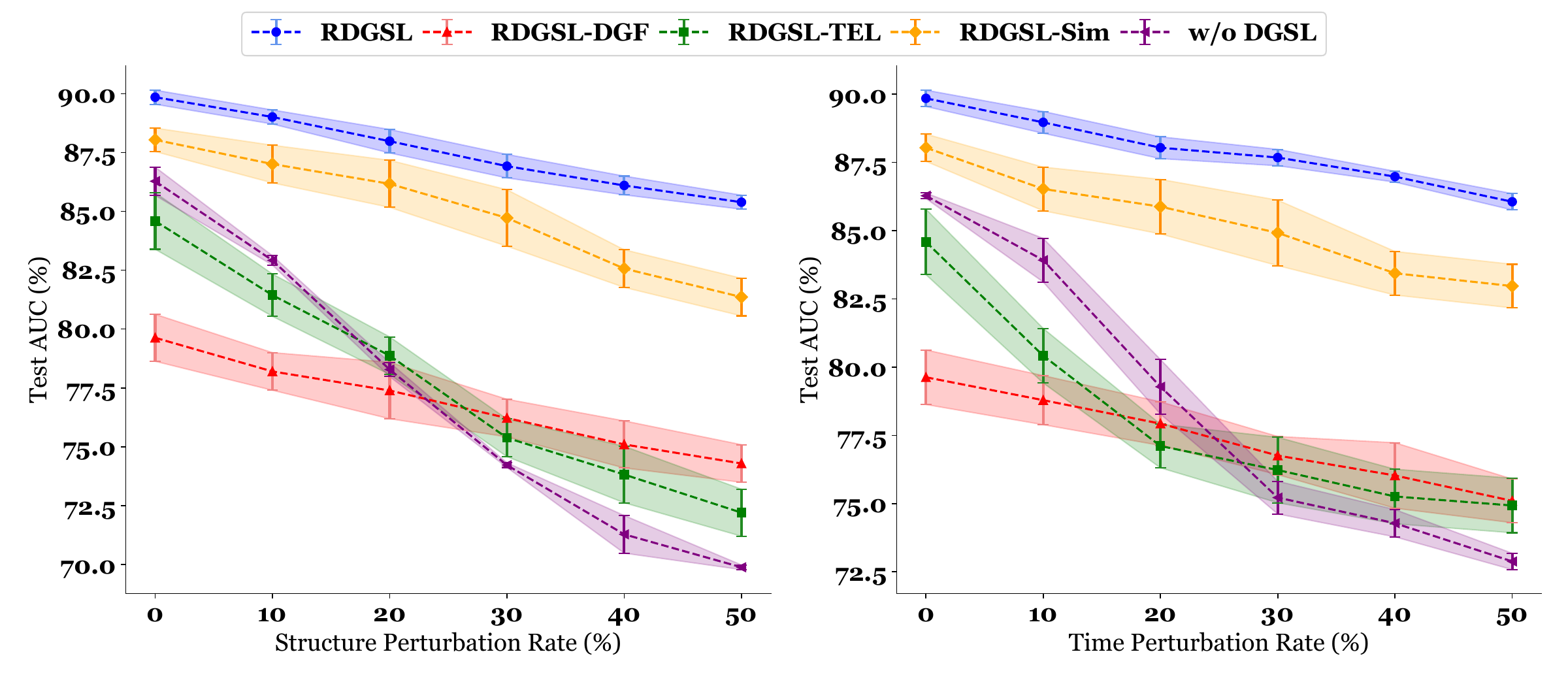}
  \caption{Ablation study on structure and time perturbation.}
  \label{ablation}
\end{figure}

The test AUC results under different levels of structure perturbation and time perturbation on Wikipedia are presented in Figure~\ref{ablation}. Remarkably, RDGSL achieves the highest performance when utilizing all components, and the performance declines when each component is removed or replaced with the existing ones. Notably, RDGSL-Sim exhibits inferior performance compared to RDGSL. This outcome substantiates the effectiveness of our proposed dynamic noise function, which dynamically captures the temporal aspect of noise in dynamic graphs. Furthermore, as the perturbation rate increases, RDGSL remains stable, whereas w/o DGSL experiences a significant decline, further validating the robustness and effectiveness of dynamic graph structure learning.

\subsection{Sensitivity Analysis}
We perform comprehensive experiments to investigate the key hyper-parameters in RDGSL. Concretely, we explore the impact of four important hyper-parameters: $\gamma$ that controls dynamic graph structure learning (Equation \ref{lossFinal}), $Q$ that presents the number of negative samples (Equation \ref{lossDGC}), $h$ that denotes the number of sampled neighbors (Section \ref{sec:TemporalSim}), and $\epsilon$ that controls the influence of dynamic noise function (Equation \ref{lossDGC}). We conduct these experiments on the evolving node classification task using the original graph and structure perturbation on Wikipedia. The results are shown in Figure \ref{sensitivity}. We observe that a larger value of $\gamma$ leads to higher AUC in original graphs, but the trend is opposite in perturbation graphs. We speculate this may be caused by noise levels in graphs: the noise in perturbation graphs is more severe, which could diminish our method's noise reduction capability under high $\gamma$. Moreover, increasing the number of negative neighbors ($Q$) enhances the performance of our method, and the contributions of $\epsilon$ and $h$ to RDGSL exhibit an initial increase and then a decrease. 
\begin{figure}
  \centering
  \begin{subfigure}{\linewidth}
    \centering
    \includegraphics[width=.9\linewidth]{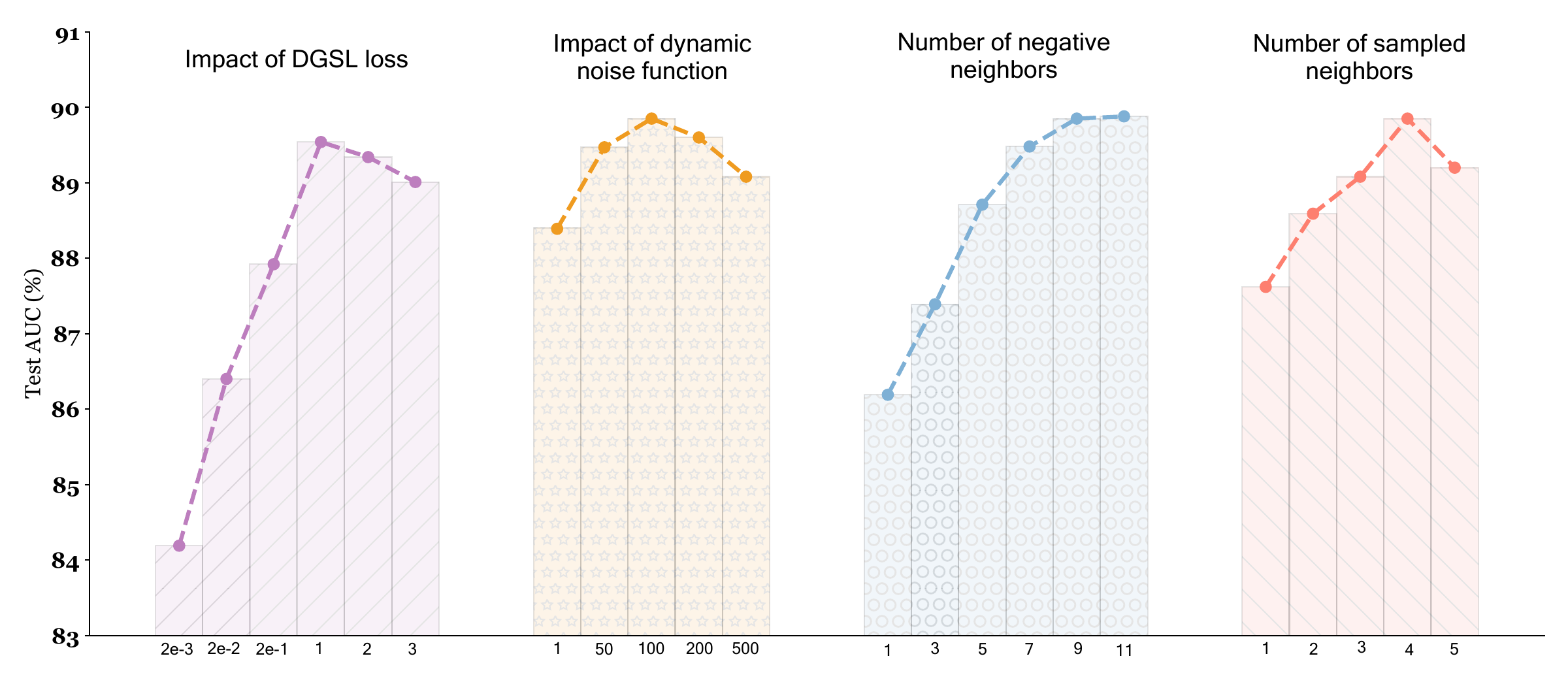}
    \caption{Sensitivity analysis on the original graph.}
    \label{gamma}
  \end{subfigure}
  \centering
  \begin{subfigure}{\linewidth}
    \centering
    \includegraphics[width=.9\linewidth]{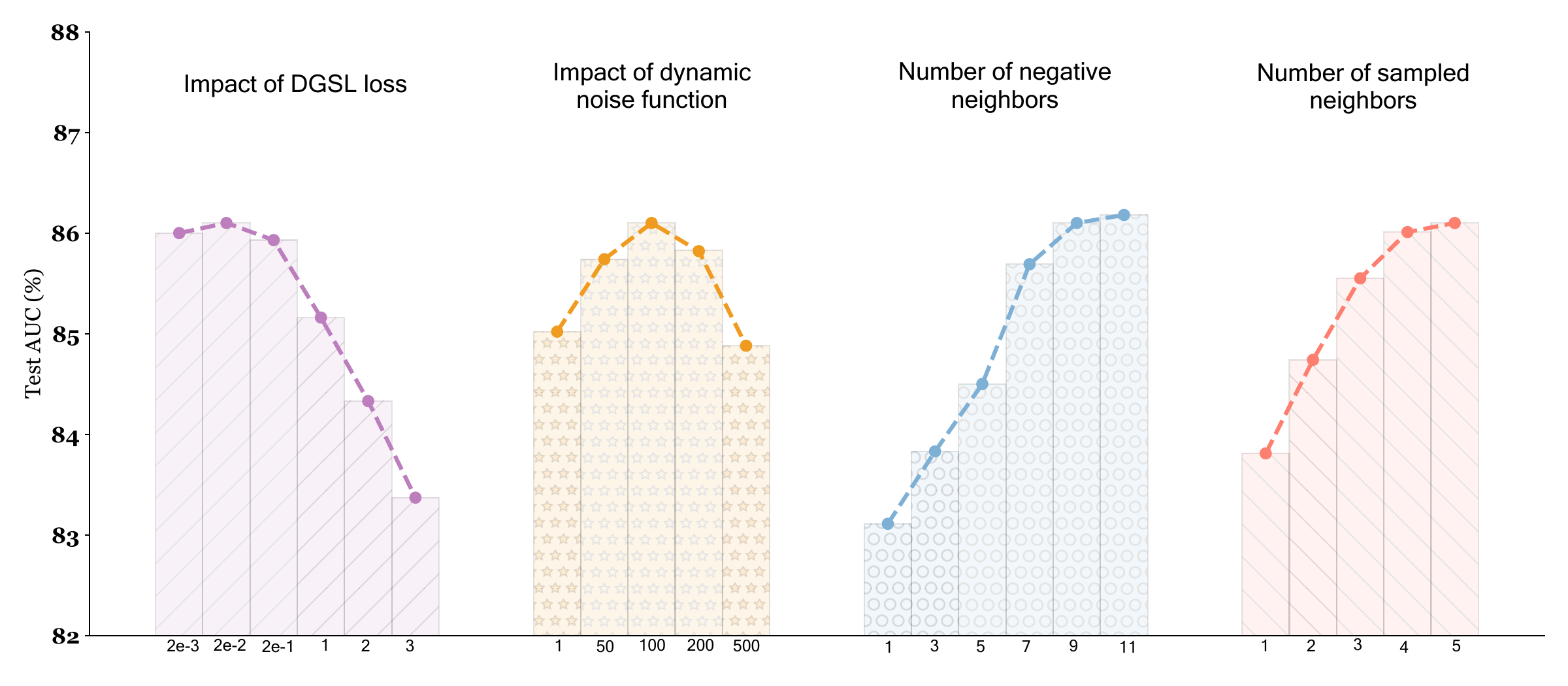}
    \caption{Sensitivity analysis with the structure perturbation.}
    \label{epsilon}
  \end{subfigure}
  \caption{Sensitivity analysis on Wikipedia.}
  \label{sensitivity}
\end{figure}

\section{Conclusion and future work}
In this paper, we present RDGSL, a concrete dynamic graph representation learning method with structure learning. We further propose dynamic graph structure learning, a supervisory signal tailored for dynamic graphs that equips our method with denoising ability. Our dynamic noise function is able to dynamically capture the temporal noise and then generate a denoised graph. Meanwhile, our attention mechanism can generate representation that remains resilient to noise, increasing the expressiveness of our method in noisy dynamic graphs. For future work, the noise we assumed in this paper is mainly in edges, while studying the noise in nodes' attributes is also one of the most vital directions in the future.


\begin{acks}
This work is funded in part by the National Natural Science Foundation of China Projects No. U1936213, No. 62206059, the Shanghai Science and Technology Development Fund No.22dz1200704, China Postdoctoral Science Foundation 2022M710747, and also supported by CNKLSTISS.
\end{acks}

\clearpage

\bibliographystyle{ACM-Reference-Format}
\balance
\bibliography{sample-base}


%\subsection{Results of link prediction}
%The results of the link prediction task on noisy graphs in Section \ref{sec:linkprediction} are shown in Table \ref{linkPrediction}.

\end{document}